%% file: main.tex
\documentclass[times,twoside]{SilvaLab_arxiv_2col}

\usepackage{amsmath}
\usepackage{amssymb}
\usepackage{amsthm}
\usepackage{amsfonts}
\usepackage{verbatim}
\usepackage{authblk}
\usepackage{graphicx}
\usepackage{comment}

\input{math_commands.txt}

\renewcommand\thesection{\arabic{section}}
\renewcommand\thesubsection{\thesection.\arabic{subsection}}
\leadauthor{George} 

\begin{document}

\title{Learning without gradient descent encoded by the dynamics of a neurobiological model}
\shorttitle{Learning without gradient decent}

\author[1,2]{Vivek Kurien George}
\author[1,2]{Vikash Morar}
\author[3]{Weiwei Yang}
\author[2]{Jonathan Larson}
\author[2]{Bryan Tower}
\author[2]{Shweti Mahajan}
\author[1,2]{Arkin Gupta}
\author[3]{Christopher White}
\author[1,2,4 \Letter]{Gabriel A. Silva}

\affil[1]{Department of Bioengineering, University of California San Diego, La Jolla CA 92093 USA}
\affil[2]{Center for Engineered Natural Intelligence, University of California San Diego, La Jolla CA 92093 USA}
\affil[3]{Microsoft Research, Redmond WA 98052 USA}
\affil[4]{Department of Neurosciences, University of California San Diego, La Jolla CA 92093 USA}

\maketitle

%TC:break Abstract
%the command above serves to have a word count for the abstract
\begin{abstract}
\noindent The success of state-of-the-art machine learning is essentially all based on different variations of gradient descent algorithms that minimize some version of a cost or loss function. A fundamental limitation, however, is the need to train these systems in either supervised or unsupervised ways by exposing them to typically large numbers of training examples. Here, we introduce a fundamentally novel conceptual approach to machine learning that takes advantage of a neurobiologically derived model of dynamic signaling, constrained by the geometric structure of a network. We show that MNIST images can be uniquely encoded and classified by the dynamics of geometric networks with nearly state-of-the-art accuracy in an unsupervised way, and without the need for any training.
\end {abstract}
%TC:break main
%the command above serves to have a word count for the abstract


\begin{corrauthor}
GS (gsilva\at ucsd.edu)
\end{corrauthor}

\section{Introduction}
In general, the tremendous success and achievements of the many flavors of machine learning (ML) are based on variations of gradient descent algorithms that minimize some version of a cost or loss function (\cite{Zhang_Bengio_Hardt_Recht_Vinyals_2016, Ruder_2016}. At their core, existing algorithms take advantage of the stochastic convergence of weights in neural networks, with individual nodes typically expressed as some version of the classical perceptron model, and the network's ability to capture latent non-trivial statistical associations that encode inputs. A fundamental limitation, however, is the need to train these systems in either supervised or unsupervised ways by exposing them to large numbers of training examples. In some situations these limitations pose significant issues, such as when there simply is not sufficient data (or enough high quality data) for training, or when tasks such as learning and classification need to be done `on the fly' in near real time to support just-in-time inference or decision making. In addition to requiring large amounts of labeled data for training, state-of-the-art ML models such as GPT3, at 175 billion parameters (\cite{Brown_2020}), require a huge compute infrastructure. The compute cost of the GPT3 training cycle alone is estimated to be \$4.6M (https://lambdalabs.com/blog/demystifying-gpt-3/). Current large data, large compute, and large model trends in ML will not scale. And while some existing ML can perform near real time learning, it still requires expensive pre-trained models. Clearly a more data, compute and energy efficient paradigm for ML is needed. 

In this paper, we introduce a fundamentally novel conceptual approach to ML that begins to address these limitations. We take advantage of a recent construction and theoretical framework - the competitive-refractory dynamics model (CRDM)- derived from an abstraction and analysis of the canonical neurophysiological principles of spatial and temporal summation. We show that when artificial neural networks (ANN) are constructed with a defined geometric and connectivity structure the interplay between this structure and dynamic variables (conduction velocities and node refractory states) allow information (inputs) to be encoded by the resultant dynamics of the network. Learning, in the traditional sense of adjusting weights, can still occur but in a much more efficient manner. No \emph{a priori} training of the network is required. The dynamics capture the characteristics of the inputs. As a proof of concept, we use these methods for unsupervised classification of MNIST digits. An expanded follow up paper will discuss the approach and methods in more detail, and provide a number of additional results.

\section{Theoretical Model}
Conceptually, the framework models the competing interactions of signals incident on a target downstream node (e.g. a neuron) along directed edges coming from other upstream nodes that connect into it. The model takes into account how temporal latencies produce offsets in the timing of the summation of incoming discrete events due to the physical geometry of the edges, in addition to the connectivity, and how this results in the activation of the target node. It captures how the timing of different signals compete to ‘activate’ nodes they connect into. At the core of the model is the notion of a refractory  state for each node. This reflects a period of internal processing at the individual node level. The model does not assume anything about the internal dynamics that produce this refractory state. Given these results, an extension of the classical model of the perceptron was achieved, a geometric dynamic perceptron that is a generalization of integrate and fire models in neuroscience. This model incorporates a timing constraint to the summation of arriving signals and subsequent edge weights as a function of edge path lengths and the neuron's refractory period. We refer the reader to \cite{10.1162/neco_a_01241} for the full details and mathematical proofs. And to \cite{Buibas_Silva_2010} and \cite{Silva_Muotri_White_2020} for additional related work. 

In the results and discussion that follow, we take advantage of the activation paths generated by the dynamics of the model, induced by inputs, i.e. activated pixels from MNIST digits. The resultant computed paths are used with or without spike-timing-dependent plasticity (STDP) (\cite{gerstner1996neuronal, ormond2009disinhibition}) directly (sections \ref{sec:N2V} and \ref{sec:N2VSTDP}) or with STDP on the weights (section \ref{sec:STDPweights}) to encode features for back end classifiers.

\section{Unsupervised Classification from Structural Paths Derived from Network Dynamics with No Training}
\subsection{Node2Vec-like GNN Implementation with no learning Rule} \label{sec:N2V}
\begin{figure}[ht]
\centering
\includegraphics[scale=0.47]{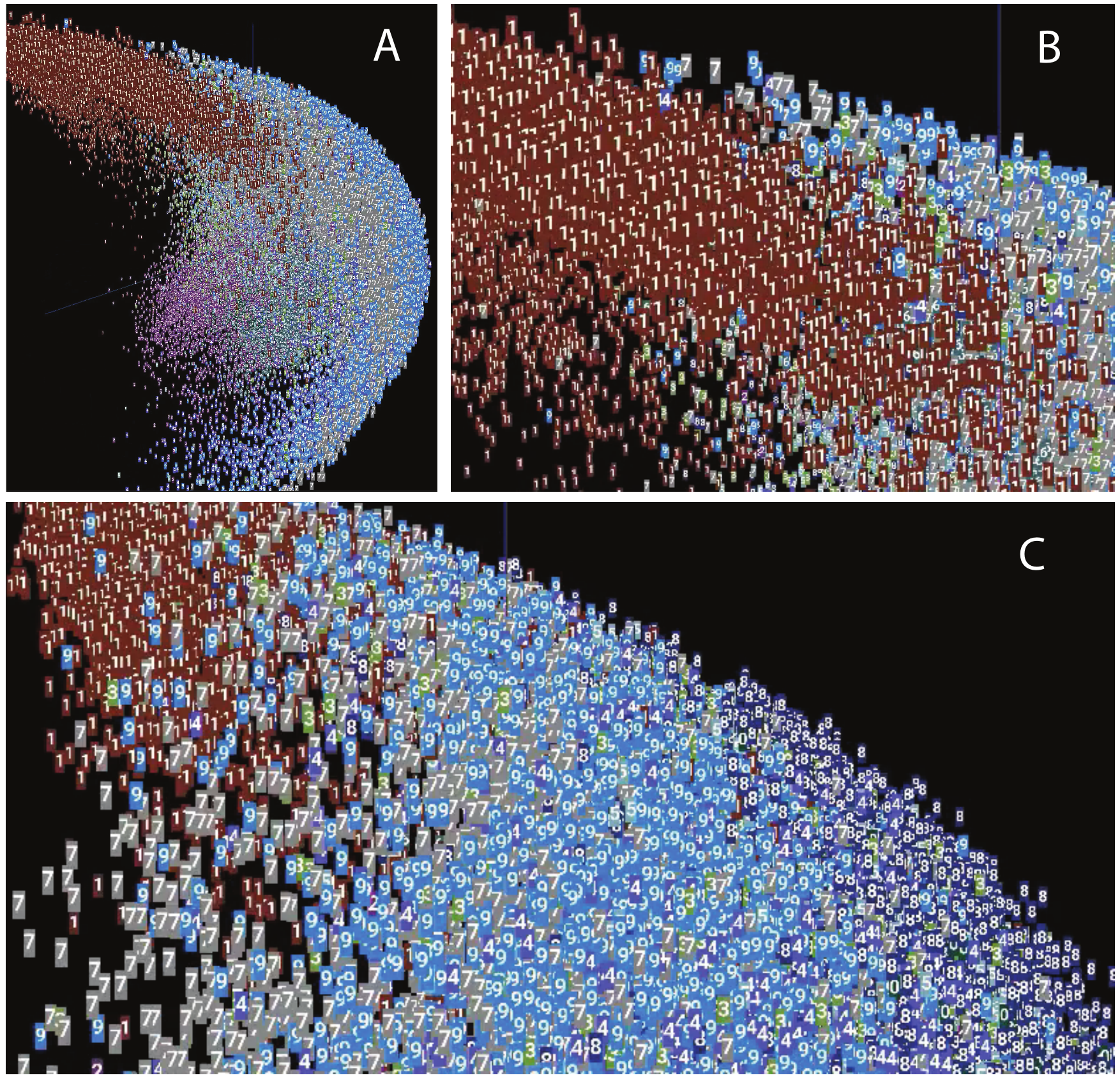}
\caption{Visualization of the first 3 PCA components of MNIST temporal graph embeddings. \textbf{A.} The full data set visualized from an arbitrarily chosen perspective in the 3D space of the PCA component axes. \textbf{B.} Zoomed in view showing the clustering boundary of the class (digits) 1' (brown). \textbf{C.} Zoomed in view (rotated to a different angle) showing the clustering boundary of the class '9' (blue).}
\label{fig:pcafig}
\end{figure}

We first explored if patterns of activation using CRDM could effectively separate out input patterns without any associated learning. In this experiment, the underlying geometric structure of our model consisted of a network with 2 blocks from a  stochastic-block model (SBM); constructed as a 784 node input block, which matched the resolution of the MNIST data, connected to a 200 hidden node block. We applied 60,000 training examples from the MNIST dataset as input stimuli. For each stimulation we ran our model for 10,000 steps, thus generating 60,000 unique activation patterns or temporal graphs.  We then generated embeddings for these graphs using their temporal sequences (paths) (\cite{George_Puppo_Silva_2020}). We visualized each embedding of a temporal graph in Euclidean space (Fig. \ref{fig:pcafig}) by plotting the first three components of a principle component analysis (PCA). Our embedding method is similar to node2vec and graph2vec, popular graph neural network (GNN) methods, with the exception that we replaced random walk paths with temporal paths generated from our simulations. 

Using cosine distance as the distance metric, we constructed a k-nearest neighbor (kNN) classifier to infer input class labels based on the majority labels of the closest neighbors. Using this unsupervised method, we were able to achieve an average accuracy of 72\% on the task of inferring the correct MNIST input class, with class 1 achieving the highest accuracy of 93\% and class 8 with the lowest accuracy  of 63\%. 

\subsection{Node2Vec-like GNN Implementation with a Spike Timing Dependent Plasticity (STDP) Learning Rule} \label{sec:N2VSTDP}
\begin{figure}[h]
\centering
\includegraphics[scale=0.53]{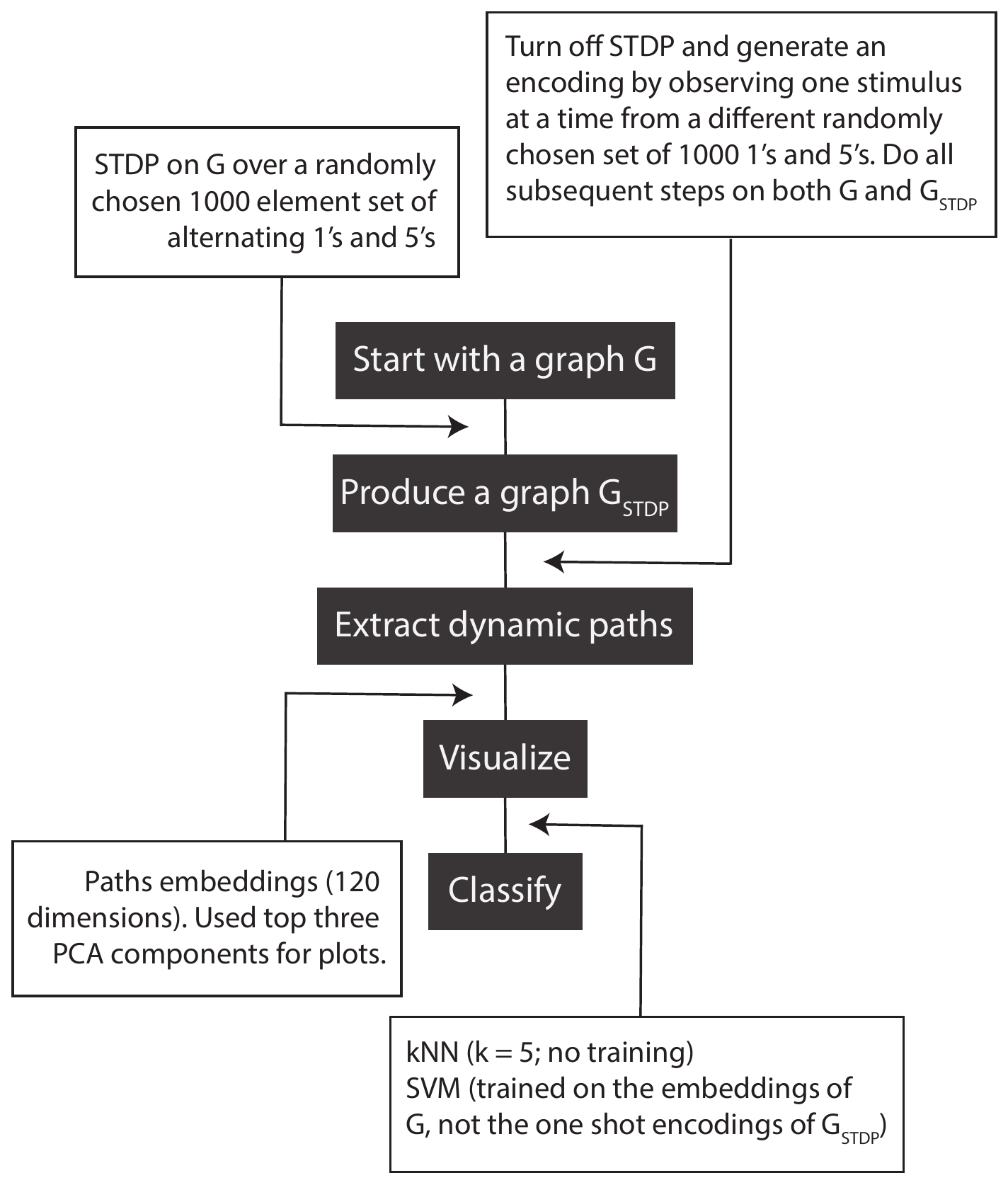}
\caption{Pipeline and workflow for STDP learning from dynamic paths.}
\label{fig:pipeline}
\end{figure}

In the next experiment, we compared embeddings generated by the CRDM with STDP as a learning rule to embeddings generated by the model without STDP (Fig. \ref{fig:pipeline}). STDP is a plausible biological algorithm that adjusts the strength of synaptic associations between connected neurons. The process adjusts synaptic weights based on the timing of arriving signaling events (see for example \cite{Lee_Panda_Srinivasan_Roy_2018}). Starting with the same underlying geometric structure of a network (\emph{c.f} section \ref{sec:N2V}), we randomly selected sets of 1000 instances of MNIST images of class 1 and 5 as input stimuli. With STDP enabled, we stimulated the model with the 1000 instances of images. We then took a snapshot of the graph with the updated weights, $G_{STDP}$. When we compared $G$ to $G_{STDP}$ we observed that 29\% of the edges had higher weights, 40\% of edges had lower weights, 20\% became inhibitory, and 11\% of the edge weights were unchanged.  We then ran two sets of simulations, both with a single additional stimulus of either class 1 or 5; one using graph $G$ and one with graph $G_{STDP}$. We then extracted the dynamic paths from these additional stimulations and embedded them using the same methods as described in section \ref{sec:N2V}. Using the same unsupervised kNN method, we inferred stimuli class labels. For embeddings generated with the dynamic paths without STDP, we achieved 61\% accuracy separating out the two input classes while for dynamic paths with STDP as a learning rule, we achieved 82\% accuracy. By comparison, if we trained a support vector machine (SVM), we were able to achieve 91.1\% accuracy with embeddings from non-STDP paths and 97.6\% for those with STDP. 

\section{Unsupervised Classification from State-Space Trajectories of Edge Weights with No Training} \label{sec:STDPweights}
In sections \ref{sec:N2V} and \ref{sec:N2VSTDP} and related published work (\cite{George_Puppo_Silva_2020,Roldan_G_George_Silva_2020}), activation patterns generate unique temporal sequences carved by the dynamics of the network that are able to classify the inputs they represent. Building on this work, we tested whether STDP-imposed changes on the weights themselves could similarly encode inputs as a function of the resultant dynamics using CRDM. In contrast to the traditional training and testing phases of an ANN, we instead observed the state-space trajectory of the evolution of the weights in the recurrent layer. Inputs of the same class of MNIST digits resulted in similar edge weight changes. Finally, we used a simple Euclidean distance weighted kNN (w-kNN, k = 5) to quantify the similarity of edge weight state-space trajectories.

To initialize the networks, we used neurobiologically relevant parameter values (\cite{Henze_2000}, \cite{Clopath_2010}). Weights were chosen from a uniform distribution, such that 70\% of the edges were excitatory while 30\% were inhibitory (\cite{gulyas1999total}).The network consisted of an input layer and a recurrent layer. We observed the resultant global dynamics, i.e. edge weight states, of the network after a single stimulation. The input layer consisted of the same 784 nodes. We varied the number of recurrently connected neurons up to 400. We did not test bigger networks because the classification accuracy did not significantly change beyond 200 nodes, although computational demands increased. 

Input nodes were activated by the non-zero pixels in a particular MNIST image. Each input node connected to every recurrent node, but the outgoing signals from each input node arrived at different times to each recurrent node due to the variability in edge delays. Individual simulations were carried out for 600 ms, a window that empirically exceeded the period of convergence to maximal classification accuracy. For each simulation, we sampled all the weights in the recurrent layer every 100 ms of simulated time. We used the resulting vectors for each time point in a w-kNN algorithm to determine the input class of specific ones. In particular, we randomly chose 9000 edge weight vectors to set up the weight space and then classified the remaining 1000 vectors using w-KNN(n=5). We did this ten times to avoid any selection bias in the weight space and for the classified vectors.

\begin{figure}[h]
\begin{center}
\includegraphics[scale=0.53]{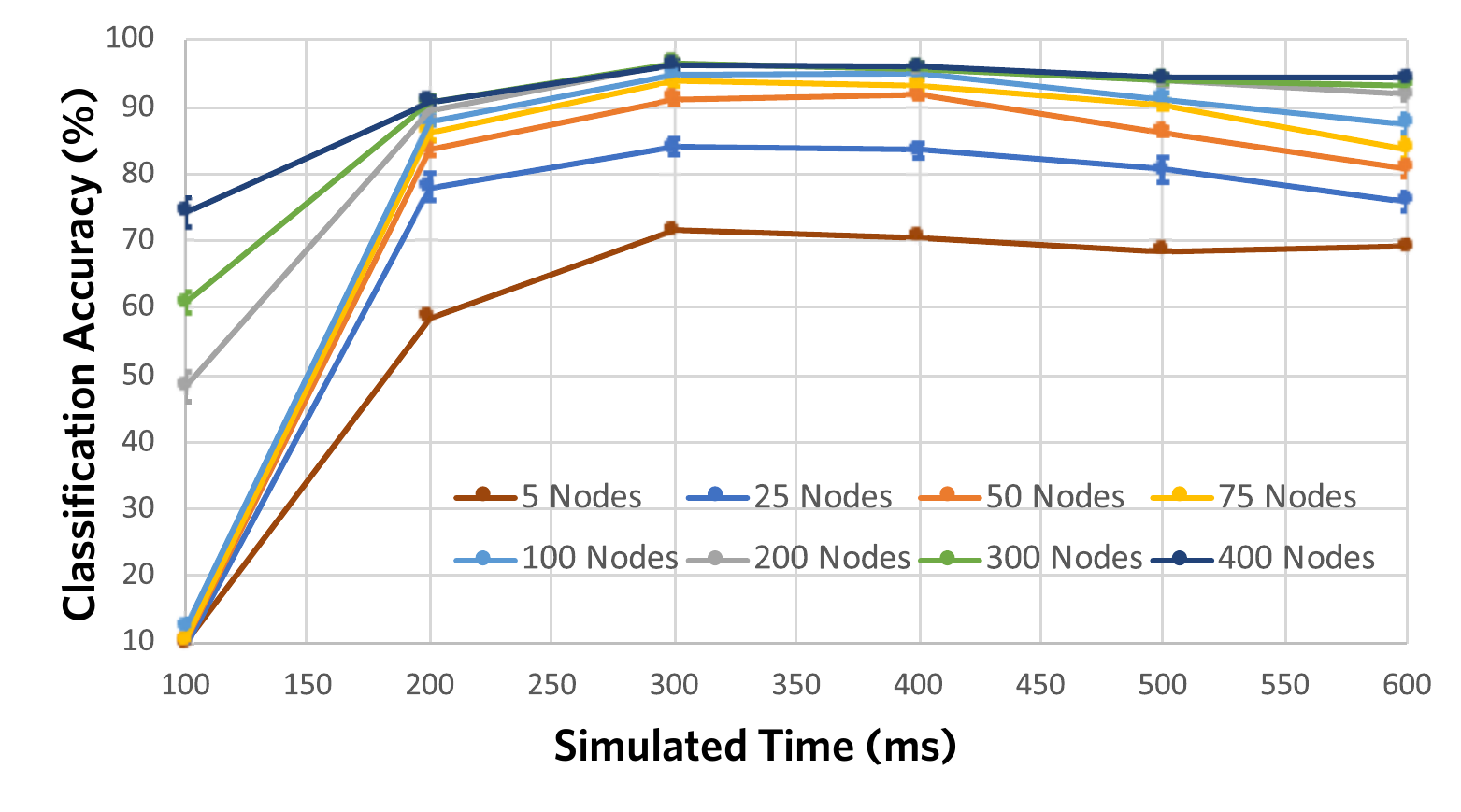}
%\framebox[4.0in]{$\;$}
%\fbox{\rule[-.5cm]{0cm}{4cm} \rule[-.5cm]{4cm}{0cm}}
\end{center}
\caption{Classification accuracy as a percentage versus simulated time in milliseconds for 10,000 instances of MNIST digits tested across networks varying from five to 400 recurrently connected nodes. See the text for full details.}
\label{fig:weightsclass}
\end{figure}

Using a euclidean w-kNN classifier metric, the model accurately predicted the correct digit 96.49\% of the time using a 200 node network after 300 ms of simulated time with no training (Fig. \ref{fig:weightsclass}).

We tested different sized networks ranging from five to 400 nodes in the recurrent layer. A five node network achieved a classification accuracy peak of 71.66\%. A 200 node network achieved the highest accuracy (96.49\%) for the most economical size. For larger networks (300 nodes and greater), the classification accuracy achieved a peak of 96.48\% that did not scale further with size of the recurrent layer. Furthermore, all the networks achieved their highest respective accuracy at about 300 ms of simulated time.  

\subsection{Comparing Edge-Weight State-Space Unsupervised Classification to a Traditional ANN}
In order to compare our approach with conventional machine learning, we benchmarked our model against an ANN implementation. We used Tensorflow to construct a sequential model with three layers: an input layer consisting of $28 \cdot 28$ nodes, a hidden layer with varying number of nodes, and an output layer with $10$ nodes. To implement one-shot learning on the ANN, we trained the network until the the model over-fitted the training data using randomly chosen images, one from each class. We used the Adam optimizer, Cross Entropy loss function and a ReLu node activation model.

\begin{figure}[h]
\begin{center}
\includegraphics[scale=0.47]{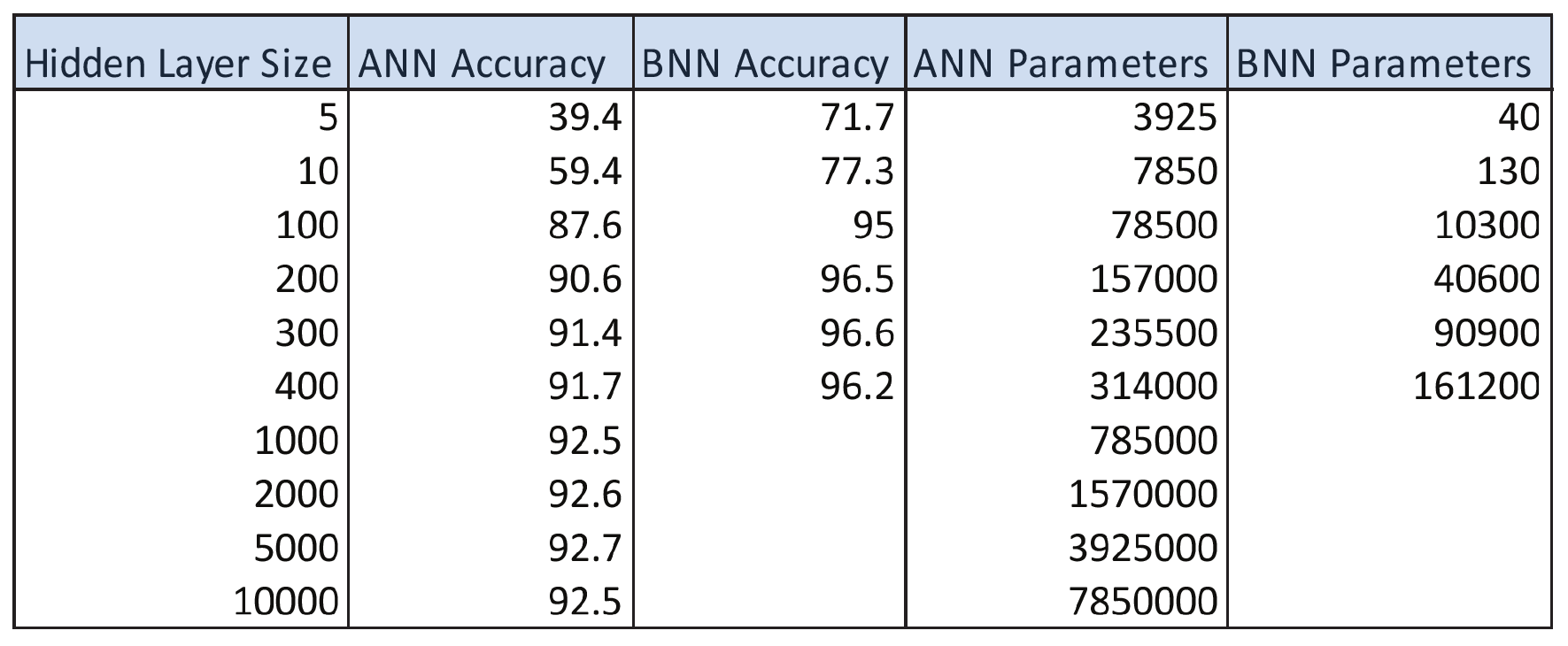}
%\framebox[4.0in]{$\;$}
%\fbox{\rule[-.5cm]{0cm}{4cm} \rule[-.5cm]{4cm}{0cm}}
\end{center}
\caption{Comparison of the ANN and our BNN (biologically-inspired neural network) with respect to classification accuracy and number of parameters. The 400 recurrent node network was the largest used in our experiments. See the text for details.}
\label{fig:table}
\end{figure}

\begin{figure}[h]
\begin{center}
\includegraphics[scale=0.57]{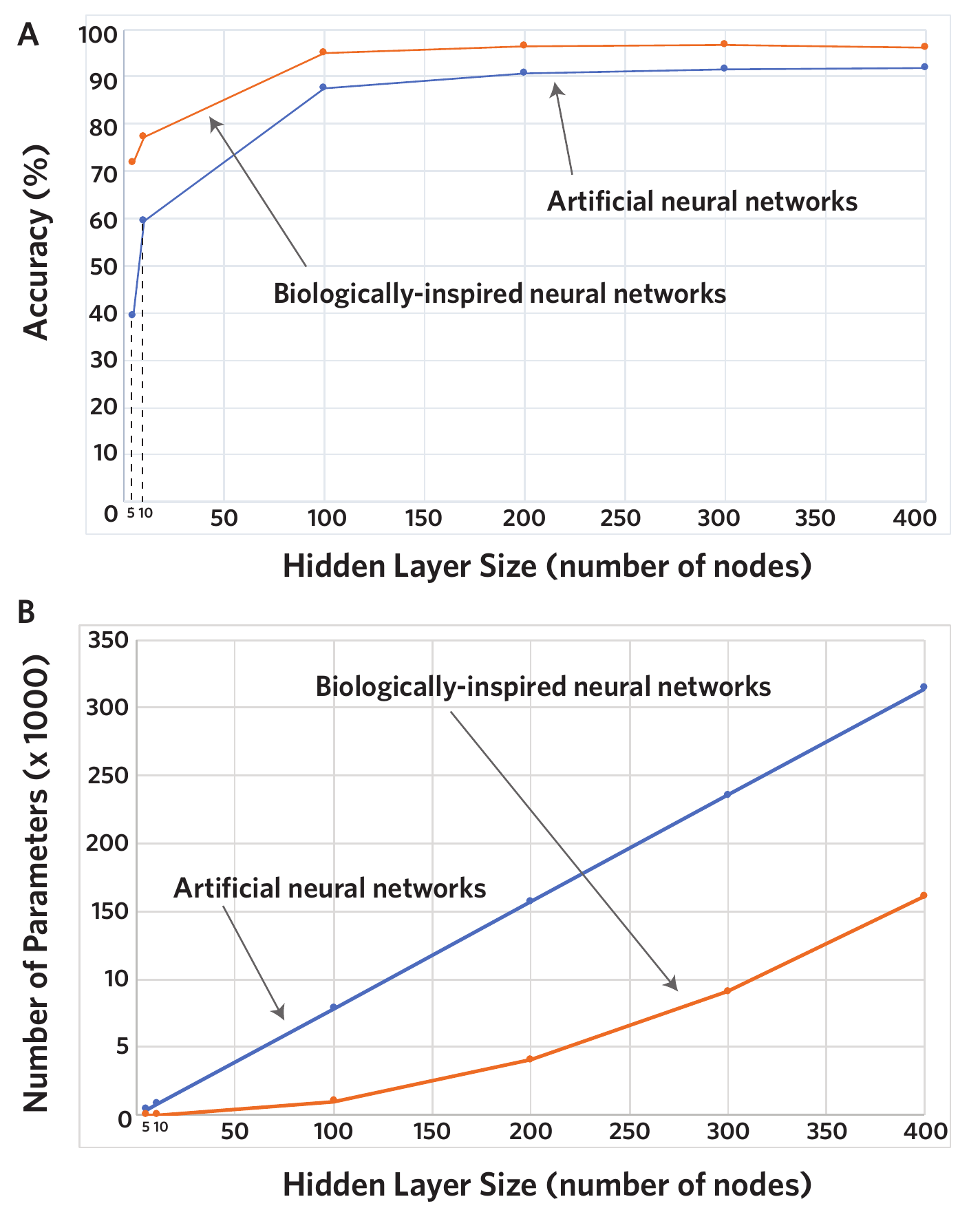}
%\framebox[4.0in]{$\;$}
%\fbox{\rule[-.5cm]{0cm}{4cm} \rule[-.5cm]{4cm}{0cm}}
\end{center}
\caption{Plot of classification accuracy (\textbf{panel A}) and number of parameters (\textbf{panel B}) as a function of the hidden layer size for the ANN and BNN models.}
\label{fig:comps}
\end{figure}

Following the one-shot training phase, we constructed an embedding space using $9000$ images from our test set. Each dimension of the embedding space corresponded to a hidden layer node. The value of each point in this embedding space was the activation values of all hidden layer nodes. After constructing the embedding space, we used a different set consisting of $1000$ test images to calculate the classification accuracy. For each of the $1000$ images, we applied a cosine distance based w-KNN. As with our CRDM spiking model above, we ran the experiment 10 times with the training and test sets redistributed each time to account for any biases in the individual selections. The data in Fig. \ref{fig:table} summarizes the numerical results. The ANN consistently performed worse on one-shot classification accuracy for each of the hidden layer sizes in comparison to our model (Fig. \ref{fig:comps}A). It also used  many more parameters than our approach with an apparent increasing divergent trend as a function of the size of the network (Fig. \ref{fig:comps}B).

\begin{comment}
\begin{table}[]
    \centering
    %\resizebox{\columnwidth}
    \begin{tabular}{|l l l l l|}
     \hline
     HL Size & ANN Acc \% & BNN Acc \% & \# ANN Pars & \# BNN Pars \\ [0.1ex]
     \hline
     5 & $39.4\pm3.5$ & $71.7\pm1.2$ & 3,925 & 40 \\ 
     \hline
     10 & $59.4\pm1.6$ & $77.3\pm1.0$ & 7,850 & 130 \\
     \hline
     100 & $87.6\pm0.4$ & $95.0\pm0.6$ & 78,500 & 10,300 \\
     \hline
     200 & $90.6\pm0.2$ & $96.5\pm0.7$ & 157,000 & 40,600 \\
     \hline
     300 & $91.4\pm0.2$ & $96.6\pm0.6$ & 235,500 & 90,900 \\
     \hline
     400 & $91.7\pm0.2$ & $96.2\pm0.8$ & 314,000 & 161,200 \\ 
     \hline
     1000 & $92.5\pm0.1$ & -- & 785,000 & -- \\
     \hline
     2000 & $92.6\pm0.1$ & -- & 1,570,000 & -- \\ 
     \hline
     5000 & $92.7\pm0.1$ & -- & 3,925,000 & -- \\
     \hline
     10000 & $92.5\pm0.1$ & -- & 7,850,000 & -- \\ [1ex] 
     \hline
    \end{tabular}
    \caption{A comparison of the ANN and our BNN (biologically-inspired neural network) with respect to classification accuracy and number of relevant parameters. The 400 recurrent node network was the largest used in our experiments due to computational costs.}
    \label{tab:1}
\end{table}
\end{comment}

\section{Discussion}
In this paper we show that images can be uniquely encoded in two different ways by the dynamics of geometric networks capable of achieving high accuracy unsupervised classification, without the need for any training of the network: Temporal sequences of activation patterns with and without STDP, and dynamical STDP mediated structural edge weight changes. To the best of our knowledge our results are the first of their kind \cite{tavanaei2019deep,lecun2015deep,schmidhuber2015deep}. 

Edge weight vectors derived from the dynamics of the model for different instantiations of the same digit class are closer in w-kNN distance to each other than members of different digit classes. This suggests that the similar images cause similar firing patterns that result in similar weight changes when STDP is applied. In other words, vectors for the same digit class are pushed through the state-space in similar directions. The resultant evolving weight state-space dynamics are sufficient to encode the latent information that characterizes the input images.

A number of open questions about how and why this approach is successful remain to be fully explored. However, the functional constraints imposed by the geometrical construction of the networks and the interplay this has given the dynamic model imposed on top of the structure appear to be the key to how information can be encoded and separated without the need for training (\emph{c.f.} \cite{Silva_Muotri_White_2020}). From a practical perspective, this fundamentally new non-gradient descent approach to machine learning opens up completely new applied directions and uses.

\section*{References}
\bibliography{main}

\end{document}

%% file: math_commands.txt
\usepackage{amsmath,amsfonts,bm}

\def\eqref#1{equation~\ref{#1}}
\def\1{\bm{1}}

\DeclareMathAlphabet{\mathsfit}{\encodingdefault}{\sfdefault}{m}{sl}
\SetMathAlphabet{\mathsfit}{bold}{\encodingdefault}{\sfdefault}{bx}{n}